# Deep Learning Model with GA-based Feature Selection and Context Integration

Ranju Mandal
*School of Engineering and Technology*
*Central Queensland University*
Brisbane Australia
r.mandal@cqu.edu.au

Basim Azam
*School of Engineering and Technology*
*Central Queensland University*
Brisbane, Australia
b.azam@cqu.edu.au

Brijesh Verma
*School of Engineering and Technology*
*Central Queensland University*
Brisbane, Australia
b.verma@cqu.edu.au

Mengjie Zhang
*Evolutionary Computation Research Group*
*School of Engineering and Computer Science*
*Victoria University of Wellington*
Wellington, New Zealand
mengjie.zhang@ecs.vuw.ac.nz

*Abstract*— Deep learning models have been very successful in computer vision and image processing applications. Since its inception, Many top-performing methods for image segmentation are based on deep CNN models. However, deep CNN models fail to integrate global and local context alongside visual features despite having complex multi-layer architectures. We propose a novel three-layered deep learning model that assiminlate or learns independently global and local contextual information alongside visual features. The novelty of the proposed model is that One-vs-All binary class-based learners are introduced to learn Genetic Algorithm (GA) optimized features in the visual layer, followed by the contextual layer that learns global and local contexts of an image, and finally the third layer integrates all the information optimally to obtain the final class label. Stanford Background and CamVid benchmark image parsing datasets were used for our model evaluation, and our model shows promising results. The empirical analysis reveals that optimized visual features with global and local contextual information play a significant role to improve accuracy and produce stable predictions comparable to state-of-the-art deep CNN models.

*Keywords— Image Parsing, Deep Learning, Genetic Algorithm, Scene Understanding, Semantic Segmentation*

## I. Introduction

The scene or image parsing is one of the core challenges in computer vision research. The main aim of image parsing is to partition the image into semantically meaningful objects and how these objects interact among themselves and to classify each object into one of the pre-defined classes. The pre-requisite of many high-impact computer vision applications such as robotic perception to enable it to interact with objects, hazard detection, augmented reality, image compression, autonomous vehicle navigation, and video surveillance [1]. The goal can be achieved by classifying each pixel in an image into predefined categories. The task of assigning accurate class labels to pixels yet remains a challenging task in natural imagery datasets. There are several variations in both the natural scenes and the occurrence of class objects. The current deep CNN-based architectures achieve promising results for pixel-wise labelling, although these architectures lack rich context representations, ultimately leaving image parsing yet an interesting and difficult problem. Many deep learning algorithms are available in the literature that are developed to address the scene parsing problem. There has been a substantial amount of work aimed at designing image parsing architectures using deep learning models due to the performance of CNN-based on benchmark datasets.

Fundamentally, image parsing is a difficult and considerably challenging problem that consists of various subtasks, which include the computation of superpixels, feature extraction from computed regions, training classifiers to obtain confidence scores, computation of pixel-wise relationship information, and the incorporation of visual and context-adaptive features for final prediction models. Current deep learning models with a single pipeline have been evolving and are being deployed for multiple vision tasks, by averting the traditional approaches. These developments have influenced various fields including robot navigation, and brain-computer interactions [2].

Recently proposed deep learning architectures have successfully computed pixel-wise labels for semantic segmentation tasks, however, due to the lack of specific contextual information most frameworks fail to achieve their full potential, and scene parsing tasks are challenging till now. The idea of predictions using the visual features is limited by inconsideration of contextual properties of objects occurring in the training samples. As the approach classifies each superpixel irrespective of the information occurring in the whole scene of the image. An example of such information can be the occurrence of roads and cars together, while a car is least likely to occur in a sea image. Some notable features that make image parsing a challenging task are the variation of information in the same class representation and often objects of different classes are made of similar materials and have similar representations.

Some of the recent works [3]–[6] have shown significant improvement in labelling tasks using CNN and the contextual information varying levels. CNNs hierarchically computes high dimensional features by downsampling the images, thus losing considerable information. The proposed model makes use of layered architecture however retains both the contextual and visual information to parse the image into segments. The features computed may prove redundant and produce misclassification, so designing a genetic algorithm to keep high-level attributes from the feature set, is ultimately removing the unwanted attributes. The selection of informative features is of great importance for the accuracy of classification [7].

Context plays an important role in complex image parsing by capturing the vital neighbourhood information, and this information can efficiently be used to produce class labels. The two essential types of contextual properties are the adjacent occurrences of pixels and the information derived from various classes occurring in an image. Although, modern deep learning models have achieved great accuracy but lack the integration of such contextual information to predict class labels.

The proposed model introduces and integrates a GA-selected subset of features computed using visual attributes and the relationship between superpixels present in the image to produce probability values for class labels. These values are provided to the integration layer, which combines sets of class-wise probabilities obtained from the classification of superpixels and relationship-aware contextual information.

The primary contributions of this paper can be summarised as follows:

a) We propose a new deep learning model based on GA-optimised visual features and the relationship between superpixel labels to produce a probability map. To improve visual features, we introduce one vs. all classifiers to produce the probability of labels for superpixels.

b) The model learns the relationship between object classes from the training samples. The proposed relationship-aware layer preserves the adjacent and block-wise spatial information of objects among superpixels.

c) A genetic algorithm is introduced in a novel way to select an optimum number of feature subsets to train the classifiers that help improving accuracy and to reduce the computational cost. Also, the integration layer fuses the obtained probabilities in a novel way to produce final class labels.

d) Experiments are conducted on benchmark datasets such as the Stanford Background [8] and the CamVid [9] datasets, and results are compared with state-of-the-art methods. We achieved new state-of-the-art performances on the Stanford Background Dataset (92.4, 92.2% global, and class accuracy, respectively), and 80.4% mIoU on the CamVid dataset.

The paper is organized as follows: Section II provides a brief overview of the existing works on image parsing-related tasks. Section III presents the details of the proposed model including training and testing algorithms. The experimental results are presented in section IV and section V concludes the paper.

## II. RELATED WORKS

A comprehensive up to date review of the recently proposed literature has been presented in this related work section. We explore a wide spectrum of pioneering works for image parsing architecture. To get the idea about the state-of-the-arts work in image parsing, some classical methods as well as a few recently published approaches such as encoder-decoder networks, multi-scale and spatial pyramid-based architectures, pixel-labelling CNN, models based on visual attention, and adversarial settings generative models are presented. Besides, as the feature selection method in the proposed architecture is based on a Genetic Algorithm (GA), we also present few recent GA-based feature selection approaches. A critical analysis was conducted to find out the strengths and weaknesses of the deep CNN architectures and their performance on the widely used benchmark datasets. Deep CNNs have achieved remarkable performance on several advance semantic scene parsing methods.

Global context-based models: PSPNet [10] utilizes global contextual information with the help of spatial pyramid pooling and improves performance for scene parsing tasks. Compared to the traditional convolution operator, DenseASPP's [3] atrous convolution can achieve a larger receptive field size without sacrificing spatial resolution and without increasing the numbers of kernel parameters. DenseASPP encodes higher-level semantics by producing a feature map of the input size with each output neuron processing a larger receptive field. Zhang et al. [11] proposed a weakly-supervised learning approach by taking advantage of the descriptive sentences of the training data using a semantic tree-based method to find out the configurations of the training data. The architecture parses an image into a structured configuration. It consists of two networks, a CNN labels objects pixel-wise, and the hierarchical object structure and the inter-object relations both are computed using RNN. RAPNet [12] addresses importance-aware street scene parsing to incorporate the importance of diverse object classes. The Importance-Aware feature selection method selects salient features for label predictions, and the residual atrous spatial pyramid module further enhances labelling to sequentially aggregate global-to-local context information in a residual refinement approach.

Local context-based model: model RefineNet [13] utilizes features at various levels found during the down-sampling process to obtain high-resolution prediction using long-range residual connections. The chained residual pooling effectively captures the rich background context. The pixel-wise labelling with FCN frameworks has also improved with the help of dilated convolutions, multi-scale features, and refining boundaries. Multi-scale context-based models: Yu and Koltun [14] proposed a CNN module that uses dilated convolutions to aggregates multi-scale contextual information and preserves resolution for dense prediction. The property of dilated convolutions that support the exponential expansion of the receptive field without losing resolution or coverage is incorporated in the architecture. The two major components of the state-of-the-art methods comprise multi-scale context-based and deep CNN-based models. Preliminary approaches are based on assigning class labels for image pixels aided by a group of low-level visual features produced at pixel-level [15], or a patch-level around each pixel [16].

Encoder decoder-based models: A hybrid network [17] was proposed to parse an image into two regions (lane and background). It combines a convolutional encoder-decoder model and a nonparametric Hierarchical Gaussian Process (HGP) classifier. The proposed model outperforms SegNet [18] and Bayesian SegNet [19] both quantitatively and qualitatively. Nguyen et al. [17] proposed a model for pedestrian lane detection in unstructured scenes that combines a compact convolutional encoder–decoder network and a nonparametric hierarchical classifier to obtain pixel-wise multi-dimensional features from an image and the HGP module classifies pixel-wise.

Predictive scene parsing models: Spatio-Temporally Coupled Generative Adversarial Networks (STC-GAN) [20] combines a future frame generation model with a predictive scene parsing model. STC-GAN captures temporal representations by computing rich dynamic features and high-level spatial contexts via a spatiotemporal encoder in the future frame generation model. Besides, the proposed coupled structure shares weights and adaptively transform features between the two models in the adversarial training stage that transfers representations from unlabelled video data to predictive scene parsing. Zhu et al. [20] proposed a video prediction-based approach, which increases the size of training instances by integrating new training samples to enhance the performance of semantic segmentation frameworks. The video prediction models' ability to predict future frames is used to predict the next labels. To alleviate misalignments in synthesized samples, a mutual transmission strategy is proposed. The proposed boundary label adjustment method makes the training process immune to annotation noises and propagation artifacts along object boundaries.

Attention-based models: Height-driven attention networks for semantic segmentation proposed by Choi et al. [21] specifically deal with urban-scene images. It exploits the pixel-wise class distributions of classes as per the vertical position of a pixel. The proposed semantic segmentation networks incorporate distinct characteristics of urban-scene images by modelling pixel-wise class distributions that exist in the dataset. OCNet [22] achieves state-of-the-art performance on three semantic segmentation benchmark datasets such as Cityscapes, ADE20K, and LIP. The model architecture inspired by the self-attention approach consists of (a) object context map for each pixel and (b) represent the pixel by aggregating the features of all the pixels weighted by the similarities.

Hierarchical models: A few early methods on scene parsing include feature hierarchies and region proposal-based architectures [23], which take regions into account to generate class label results. However, features that capture global context perform better as pixel-level features cannot capture robust statistics about the appearance of a local region, and patch-wise features are prone to noise from background objects. Xiong et al. [28] used deformable ConvNets to learn adaptive feature maps in a structured fashion indoor scene parsing. In contrast to conventional ConvNets, sharing the size of spatial context, the deformable ConvNet acquires context with help of depth information. This information gives important clues to identify real local neighbourhoods. The model can aggregate flexible spatial context by expanding the acquired information kernel with usual convolution filters.

Panoptic segmentation: Panoptic segmentation task unifies instance and semantic segmentation to produce a more accurate outcome for thing classes as instance segmentation (objects with instance-level annotation) alongside for stuff classes (regions of similar textures such as roads, sky, etc.) as semantic segmentation. UPSNet [25] a single backbone residual network followed by deformable convolution-based semantic segmentation head and mask R-CNN [26] style segmentation head, resolving subtask simultaneously, and finally, a panoptic head solves the panoptic segmentation using pixel-wise classification. DeeperLab [27] performs a panoptic segmentation or whole image parsing with a simpler, fully convolutional approach that jointly addresses the semantic segmentation and instance segmentation in a single-shot, bottom-up approach. An encoder based on depth-wise separable convolution was built for faster inference, and the backbone is additionally augmented with the ASPP module for Atrous convolutions. The decoder maps detailed object boundaries and uses two large kernels (7 × 7) depth-wise convolutions to further increase the receptive field. The resultant feature map is then reduced by depth-to-space which is used as the input for the image parsing prediction heads.

Genetic algorithm-based models: GAs and other optimisation algorithms have been applied in wide range of applications. Fan et al. [29] proposed a GA-based algorithm for facial feature selection. GAs proved to be efficient in another application [30] to find the optimal solution on high-dimensional search space due to its search capabilities. Feature selection based on tribe competition-based GA divides the population into multiple tribes, and the evolutionary process ensures that the number of selected features in each group follows a Gaussian distribution. The competition-based GA p-ermits the superior tribes to have more individuals to search their parts of the solution space. Another modified version of GA [31] named Fast Rival GA (FRGA) was proposed to partition the chromosomes into winners and losers by integrating the competition concept. Some researchers have used GAs/EAs for optimising various architectures. Rahman and Verma [32] used GAs for ensemble architecture optimisation. Huang and Wang [33] used GAs for optimising support vector machines. Researchers also used evolutionary algorithms [34, 35] and genetic programming [36] to find suitable features and architectures. The feature selection mechanism at the first layer of our model is based on FRGA.

Context information is of vital importance for image parsing applications by considering the neighbouring superpixel information, and this information helps produce final class labels. The abundance of features computed makes the model heavy and complex, thus the selection of a subset of features is important for various reasons including the generalized performance, decreased computational overhead, and identification of highly important attributes. Towards this end, we propose an efficient image parsing model that introduces a genetic algorithm to compute an efficient subset of features to train the classifiers and the relationship-aware matrices to consider context information, to assign class labels to each superpixel.

III. PROPOSED DEEP LEARNING MODEL

The proposed deep learning model initially computes superpixels level visual features. The superpixels features are fed into a visual feature prediction layer which builds the class-semantic supervised classifiers. Here, the superpixels-level visual features are computed first, and these visual features are predicted using the class-semantic supervised classifiers to obtain class-wise probabilities of all superpixels. We obtain a class-wise probability vector after the completion of this step. The class probabilities of all superpixels based on their visual features are predicted using the class-semantic supervised classifiers. The contextual voting layer obtains local and global contextual features of each superpixel based on its most probable class and the corresponding Object Co-occurrence Priors (OCPs). Finally, these three features are concatenated and fed into a fully connected neural network or multilayer perceptron with a single hidden layer.

then based on this division the superpixel information matrices are computed using the training set, these matrices help produce the final class label.

*a) Adjacent Superpixel Votes*

To consider the local contextual information we make use of the adjacent superpixels, which participate to produce the probability value of the neighbourhood superpixels. Each superpixel $s_i$ votes for other superpixel $s_j$ $i \neq j$ a specific class label. A matrix is formulated for the adjacent superpixel information from the training data. For each superpixel, the

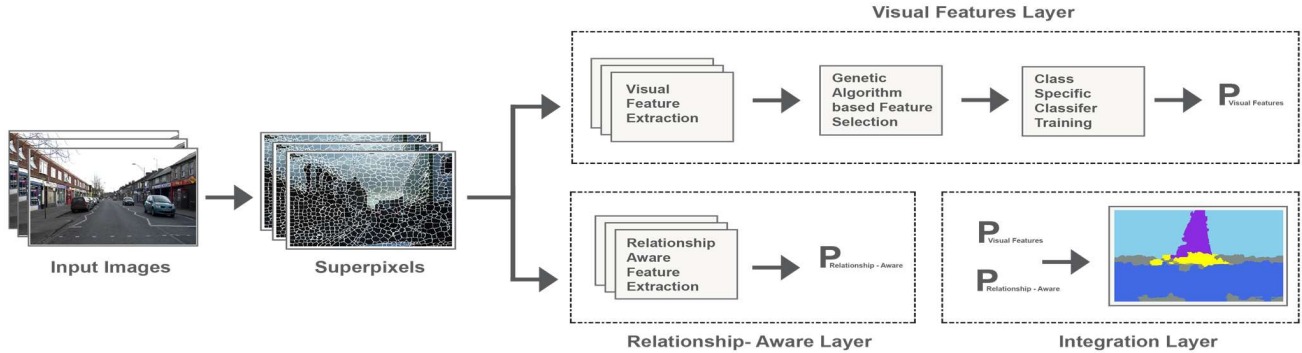

Fig. 1 Image Parsing Framework. **Visual Features Layer:** genetic algorithm computes a subset of features and individual classifiers produce the class probabilities. **Relationship-Aware Layer:** it considers votes from adjacent superpixel and from superpixel blocks, to produce probability values for class labels. **Integration Layer:** it provides final class labels for superpixels by integrating the probabilities from the visual feature layer and relationship-aware layer.

The proposed model contains several layers which make use of the contextual and visual properties of the superpixels to produce final class labels. The model layers and training (Algorithm 1) and testing (Algorithm 2) algorithms are described below: The proposed deep learning model for effective scene parsing is presented in Fig. 1.

*1) Visual Features Layer*
Initially, the superpixels are computed from the input images, these superpixels are further used to extract features, then individual classifiers are trained based on the selected features (from the feature set) for each class in the dataset. The visual features layer considers the raw superpixels computed and extract visually evident attributes to distinguish between object classes. The set of attributes computes is fed to the genetic algorithm-based feature selection algorithm [31]; the output is an optimally selected subset of highly important features. The selected subset helps training the class-wise classifiers. The visual features layer aims to produce a probability of class label for each superpixel using the trained classifiers.

*2) Relationship-Aware Layer*
In the proposed architecture we make use of the contextual properties of the object and compute probabilities based on the relationship between superpixels. These probabilities are termed as local and global, the former makes use of the adjacent superpixels values and the latter considers the superpixel classes present in a spatial block. Both the local and global information helps to produce the probability value for each superpixel. The relationship awareness is explained as initially the image $I$ is equally distributed in $k$ number of superpixels, and $B_k$ block, local superpixel information accounts for the class labels of its neighbourhood superpixels.

*b) Blockwise Superpixel Votes*

In addition to the neighborhood superpixels, the information in a block of superpixels is also used to exploit long-range dependencies. A block of superpixels is considered which votes that superpixel with class label exists in the other block.

*3) Integration Layer*

The probability values computed from the relationship-aware layer and the visual feature prediction layer are combined in the integration layer to assign class labels to each superpixel.

*4) Genetic Algorithm-based Feature Selection*

The proposed visual feature layer incorporates a genetic algorithm to select an optimum subset of features to train the classifiers. The genetic algorithm computes the subset from the total features. The number of features is of great importance to the complexity and performance of the proposed model. A large number of attributes make the model computationally expensive, and the classifiers require more time to train on a large set of features. Towards this end, the GA-based approach is designed to select the optimum and high-level features to train the classifiers. The GA uses non-overlapping populations, roulette wheel selection while for accuracy computations nearest neighbour type classifier is used. The number of chromosomes was 10 and we run the algorithm for 1000 generations. The crossover probability used is equal to 0.8, and

the mutation probability is 0.1. The fitness function is described as:

$$Fitness(y) = \alpha * error + \beta * \frac{S}{T}$$

where $y$ is the subset of optimal features, $\alpha = 0.99$, and $\beta = 0.01$ were selected based on experimentation, $S$ is the number of features selected, and $T$ is the total number of features.

---

**Algorithm 1: Procedure to train the proposed model**

| Input | : Training Data, Number of Superpixels to Compute |
|---|---|
| Output | : Trained Classifiers $\hbar_i$, Relationship-Aware Matrices $\mathfrak{m}_1, \mathfrak{m}_2$ |

**Foreach** Image I
  Compute $N$ Superpixels $S = \{s_j | j = 1,2,3, \dots, N\}$
  **Foreach** Superpixel $s_j \in S$
      Extract $m$ Visual Features
      Select Optimum Subset of Features $n$ via Genetic Algorithm
      Train Class-Specific Classifier $\hbar_i$, where $i = 1: n\_classes$
  **End**
**End**

**Foreach** Image I
  Compute $N$ Superpixels $S = \{s_j | j = 1,2,3, \dots, N\}$
  **Foreach** Superpixel $s_j \in S$
      Compute votes from Adjacent Superpixels for probable class
      Blockwise votes for the probable class of each Superpixel
      Formulate Superpixel Relationship Information Matrices $\mathfrak{m}_1, \mathfrak{m}_2$
  **End**
**End**

---

## IV. RESULT AND DISCUSSIONS

The performance evaluation of the proposed deep learning model is presented in this section. Two publicly available benchmark datasets are used for our experiments. A detailed comparative analysis of results was conducted with results obtained by the recently published state-of-the-art methods on scene parsing. Algorithm 1 describes the training procedure of the architecture while Algorithm 2 presents the testing procedure of the proposed image parsing framework.

### A. Dataset

Stanford Background Dataset (SBD) was introduced by Gould et al. [8] for evaluating methods for geometric and semantic scene understanding. The dataset comprises 715 images chosen from public datasets: LabelMe, MSRC, PASCAL VOC, and geometric context. There are eight object classes, including Sky, Tree, Road, Grass, Water, Building, Mountain, and Foreground object. The selection criteria were for the images of outdoor scenes, having approximately 320-by-240 pixels, containing at least one foreground object, and having the horizon position within the image. Semantic and geometric labels were obtained using Amazon's Mechanical Turk (AMT). All image pixels are manually annotated into one of eight classes or unknown objects.

---

**Algorithm 2: Procedure to test the proposed model**

| Input | : Test Data, Trained Classifiers $\hbar_i$, Relationship-Aware Matrices $\mathfrak{m}_1, \mathfrak{m}_2$ |
|---|---|
| Output | : Class Labels |

**Foreach** Image I
  Compute $N$ Superpixels $S = \{s_j | j = 1,2,3, \dots, N\}$
  **Foreach** Superpixel $s_j \in S$
      Predict the Probability $P_{Visual\ Features}$ of Class Label using trained Classifier $\hbar_i$
  **End**
**End**

**Foreach** Image I
  Compute $N$ Superpixels $S = \{s_j | j = 1,2,3, \dots, N\}$
  **Foreach** Superpixel $s_j \in S$
      Predict the Probability $P_{Relationship-Aware}$ for Class Label using the Matrices $\mathfrak{m}_1, \mathfrak{m}_2$
  **End**
**End**

**Foreach** Image I
  **Foreach** Superpixel $s_j \in S$
      Assign Class Label to each superpixel using $P_{Visual\ Features}$ and $P_{Relationship-Aware}$
  **End**
**End**

---

Region label files contain an integer matrix indicating each pixel's semantic class (sky, tree, road, grass, water, building, mountain, or foreground object) and a negative number indicates unknown classes. 572 images are randomly selected for training and 143 images for testing to obtain classification accuracy.

The CamVid dataset [9] is a set of videos with object class semantic labels. The database consists of ground truth labels that represent semantic classes for each pixel. The dataset provides experimental data to evaluate evolving algorithms. Most videos are acquired using fixed-position CCTV-style cameras, this data was acquired from a driving automobile perspective. The driving scenario adds diversity to the observed object classes. The per-pixel semantic segmentation of over 700 images was assigned manually and was cross-checked. The dataset's resolution was decreased to 480 × 360 from the original 960 × 720, for experiments to follow previous works. We encapsulate the pixel label data and the label ID to a class name mapping for our training and testing stages. Usually, the classes in a dataset have an equal number of observations. However, the classes in CamVid are imbalanced, a common problem in street scene imagery. Figure 2 shows the class-wise pixel imbalance in bar charts. The CamVid dataset contains more sky, building, and road pixels than pedestrian and bicyclist pixels because the sky, buildings, and roads cover more areas in the image. To mitigate this problem and just to have more training samples class-wise, 32 original classes in CamVid are grouped into 11 classes (i.e., Sky, Building, Pole, Road, Pavement, Tree, Sign-symbol, Fence, Car, Pedestrian, Bicyclist), and the dataset was divided into 4:1 ratio for training and testing respectively, for our experiments.

### B. Training Details

We implement our architecture with the public open-source Python 3 library. The experiments are conducted on an HPC cluster with 20 computer nodes, 40 CPU sockets, 528 cores, and multiple GPUs. For training, we uniformly resize each input image into 256*256*3 pixels. Due to the limited memory, the mini-batch size is set to 4. The learning rate is initialized to be $10^{-4}$, and decays exponentially with the rate of 0.1 after 30 epochs.

### C. Evaluation of the Stanford Background dataset

Table I shows the performance comparison with previous approaches on the Stanford Background Dataset. The class-wise accuracy we have obtained on the Stanford dataset and the class-wise balanced results across 8 object classes are presented. We observed an accuracy improvement of 3.02% and 0.86% on the Stanford Background and the CamVid datasets, respectively, using GA-based feature selection. Table II shows the confusion matrix we obtained on the Stanford dataset. The proposed deep learning model achieves accuracies of 92.4% (Table III) on the Stanford Background dataset, and the accuracy is comparable to the state-of-the-art methods. The accuracy of the proposed model compared with the accuracies reported by recent top-performing models is presented in Table III.

TABLE I. PERFORMANCE (%) COMPARISON WITH PREVIOUS APPROACHES ON THE STANFORD BACKGROUND DATASET

| Method | Global | Avg | Sky | Tree | Road | Grass | Water | Bldg. | Mtn. | Fgnd. |
|---|---|---|---|---|---|---|---|---|---|---|
| Gould et al. [3] | 76.4 | 65.5 | 92.6 | 61.4 | 89.6 | 82.4 | 47.9 | 82.4 | 13.8 | 53.7 |
| Munoz et al. [7] | 76.9 | 66.2 | 91.6 | 66.3 | 86.7 | 83.0 | 59.8 | 78.4 | 5.0 | 63.5 |
| Ladicky et al. [24] | 80.9 | 70.4 | 94.8 | 71.6 | 90.6 | 88.0 | 73.5 | 82.2 | 10.2 | 59.9 |
| Proposed Method (256) | **89.4** | **85.9** | **94.5** | **83.1** | **94.6** | **91.2** | **89.9** | **90.8** | **78.3** | **78.7** |
| Proposed Method (512) | **92.4** | **92.2** | **95.1** | **88.6** | **95.3** | **94.8** | **96.1** | **93.5** | **90.3** | **84.1** |

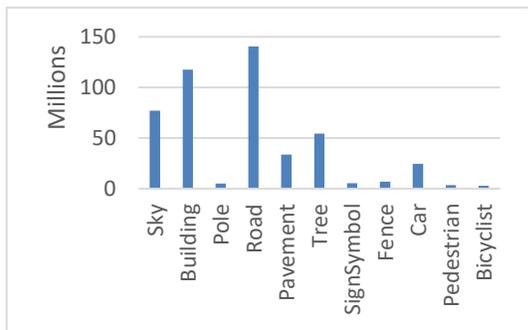

Fig. 2. Class-wise pixel distribution of 11 classes in CamVid dataset. The line chart shows that the distribution imbalanced.

TABLE III. PERFORMANCE (%) COMPARISONS WITH PREVIOUS APPROACHES ON THE STANFORD BACKGROUND DATASET

| Method | Pixel Acc. | Class Acc. |
|---|---|---|
| Gould et al. [8] (2009) | 76.4 | |
| Kumar and Koller [37] (2010) | 79.4 | - |
| Lempitsky et al. [38] (2011) | 81.9 | 72.4 |
| Farabet et al. [15] (2013) | 81.4 | 76.0 |
| Sharma et al. [16] (2015) | 82.3 | 79.1 |
| Luc et al. [39] (2016) | 75.2 | 68.7 |
| DeepLabV2 [44] (2017) | 87.0 | 75.9 |
| SegWGAN [45] (2019) | 87.7 | 79.0 |
| Proposed Method | **92.4** | **92.2** |

A class accuracy of 92.2% is observed for our network compared with 76% [44] and 79% [45] accuracies by existing methods, indicating that the proposed network tends to focus on common classes with a large proportion of training pixels. The proposed network outperforms visual feature-based

TABLE II. CONFUSION MATRIX FOR EIGHT OBJECTS ON THE STANFORD BACKGROUND DATASET (MULTILAYER PERCEPTRON (MLP), GLOBAL MEAN ACCURACY = 92.4%)

| | Sky | Tree | Road | Grass | Water | Bldng. | Mtn. | Fgnd |
|---|---|---|---|---|---|---|---|---|
| Sky | **0.95** | 0.02 | 0.00 | 0.00 | 0.00 | 0.02 | 0.00 | 0.01 |
| Tree | 0.02 | **0.88** | 0.01 | 0.01 | 0.00 | 0.12 | 0.00 | 0.04 |
| Road | 0.00 | 0.00 | **0.95** | 0.00 | 0.00 | 0.02 | 0.00 | 0.05 |
| Grass | 0.00 | 0.03 | 0.01 | **0.94** | 0.00 | 0.01 | 0.00 | 0.05 |
| Water | 0.01 | 0.00 | 0.00 | 0.01 | **0.96** | 0.01 | 0.00 | 0.15 |
| Bldng. | 0.01 | 0.06 | 0.01 | 0.00 | 0.00 | **0.93** | 0.00 | 0.04 |
| Mtn. | 0.02 | 0.01 | 0.00 | 0.03 | 0.01 | 0.01 | **0.90** | 0.04 |
| Fgnd. | 0.02 | 0.05 | 0.07 | 0.01 | 0.01 | 0.09 | 0.00 | **0.84** |

classifiers with a significant margin, confirming the contribution of the proposed architecture.

TABLE IV. PERFORMANCE (%) COMPARISONS WITH PREVIOUS APPROACHES ON THE CAMVID DATASET

| Method | Pre-trained | Encoder | mIoU (%) |
|---|---|---|---|
| Badrinarayanan et al. [18] (2019) | ImageNet | VGG16 | 60.1 |
| Huang et al. [41] (2018) | ImageNet | VGG16 | 62.5 |
| Yu and Koltun [14] (2015) | ImageNet | Dilate | 65.3 |
| Yu et al. [42] (2018) | ImageNet | ResNet18 | 68.7 |
| Zhao et al. [10] (2017) | ImageNet | ResNet50 | 69.1 |
| Bilinski and Prisacariu [43] (2018) | ImageNet | ResNeXt101 | 70.9 |
| Chandra et al. [40] (2018) | Cityscapes | ResNet101 | 75.2 |
| Proposed Approach | NA | NA | **80.42** |

### D. Evaluation of the CamVid dataset

The Intersection-Over-Union (IoU), also called the Jaccard Index, is computed to evaluate performance on the CamVid dataset. We computed the mean IoU metric (using weighted Jaccard Index) as it is mostly used for evaluation on the CamVid dataset by the top-performing methods in the literature. IoU metric is an extremely effective evaluation metric in semantic segmentation. We obtained 80.42% mIoU by the proposed model without using any pre-trained weights. A comparative evaluation on mean IoU with the state-of-art approaches on the CamVid dataset is presented in Table IV. The study shows that the proposed model has obtained a significant improvement over the existing approaches. Figure 3 presents the visual results on the Stanford Background Dataset using the proposed approach, the left column includes images from the dataset while the middle column represents the ground truth files, while the rightmost column interprets the output of the trained model.

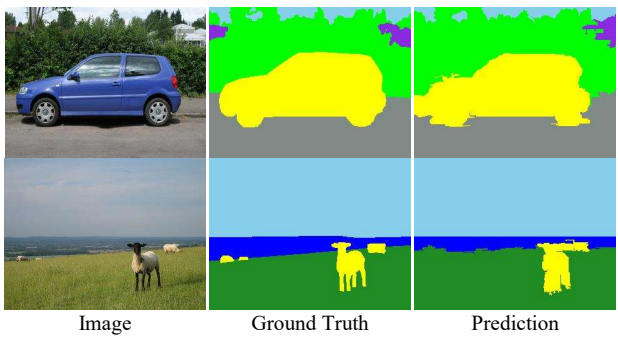

| Image | Ground Truth | Prediction |

Fig. 3. Qualitative results produced by proposed model on Stanford Background Dataset (best view in color mode). Original, annotated, and segmented images are presented column-wise.

## V. CONCLUSION

We presented a deep learning model optimized by a genetic algorithm for parsing images in complex scenarios. We computed a rich feature vector by incorporating relationship information from data and took advantage of the contextual features which captured both short- and long-range label dependencies of objects in the entire scene while being able to adapt to the local context in the scene. The GA-based optimization was investigated to find the optimized set of features to train the classifiers. The proposed deep learning model was evaluated on benchmark datasets. We achieved an accuracy of 92.4% on the Stanford Background Dataset, and 80.42% mIoU on CamVid dataset. The comparative analysis as shown in Table III and Table IV provides evidence of improved performance by our proposed model. In our future work, we will further evaluate the performance of the proposed model on large benchmark datasets. We will further improve the optimization of all layers.

## VI. ACKNOWLEDGMENT

This research was supported under Australian Research Council's Discovery Projects funding scheme (project number DP200102252).